\documentclass[letterpaper, 10 pt, conference]{ieeeconf}  

\IEEEoverridecommandlockouts                              

\usepackage{graphics} 
\usepackage{epsfig} 
\usepackage{svg} 
\usepackage{mathptmx} 
\usepackage{times} 
\usepackage{amsmath} 
\usepackage{amssymb}  
\usepackage{array}

\title{
Achieve What You Imagined: Learning to Align Actions with Visual Plans
}

\author{%
\parbox{0.9\textwidth}{\centering
Yuheng Qiao$^{1,\dagger}$~
Ziran Wei$^{1,\dagger}$~
Xiaohan Wang$^{2}$~
Daqiang Guo$^{3}$~
Yichen Luo$^{1}$~
Zhibo Pang$^{4}$\\
Peng Zhou$^{5,*}$~
Sichao Liu$^{1,*}$%
\thanks{$^{\dagger}$Equal contributor. $^{*}$Corresponding authors: Sichao Liu (sicliu@kth.se) \& Peng Zhou (pzhou@gbu.edu.cn).}%
\\[0.5em]
{\small
$^{1}$KTH~
$^{2}$Beihang University~
$^{3}$The Hong Kong University of Science and Technology (Guangzhou)~
$^{4}$Peking University~
$^{5}$Great Bay University
}%
}%
}

\usepackage{xcolor}
  
\usepackage{amsmath}
\usepackage{booktabs}
\usepackage{url}
\usepackage[colorlinks=true,linkcolor=black,citecolor=black,urlcolor=magenta]{hyperref}

\usepackage{algorithm}
\usepackage{algpseudocode}

\begin{document}
\maketitle
\thispagestyle{empty}
\pagestyle{empty}

\begin{abstract}
World-action models can jointly predict future visual observations and robot actions. However, discrepancies may exist between their visual predictions and the consequences implied by generated actions. We observe that WAMs can often generate visually plausible task-completion outcomes before producing action sequences that reliably achieve them. Consequently, we treat the WAM-generated visual prediction as a goal-conditioned visual proposal rather than a directly executable plan. We use a frozen action-conditioned world model to predict action-conditioned consequences and construct feedback based on consistency between the two future predictions and alignment with the terminal goal. Leveraging this feedback, we employ Flow Policy Optimization (FPO) to optimize the action head of the WAM. This framework avoids online robot interaction and additional training of task-specific reward models. Across four real-world UR5 manipulation tasks, our method increases the mean success rate from 43.4\% to 75.1\%, compared with 61.4\% for $ \pi_{0.5}$. These results show that cross-model prediction discrepancy can provide useful feedback for improving robot policies under the evaluated manipulation tasks. 
Website: \url{https://imagine-to-achieve.github.io/}

\end{abstract}

\section{INTRODUCTION}
Robot learning is increasingly moving from task-specific controllers toward large pretrained policies that enable transfer across tasks and robot embodiments. Vision–language–action (VLA) models map observations and instructions directly to actions~\cite{rt2,openvla,pi0}, while world–action models (WAMs) additionally predict future visual observations associated with generated actions~\cite{unipi,cosmos3}. Reinforcement learning is a common approach for post-training such policies, but its effectiveness depends on obtaining informative rollout feedback, which remains challenging in low-data settings. Binary terminal rewards are well aligned with the task but often too sparse early on: when a weak base policy fails across sampled rollouts, they provide limited task-directed signal~\cite{grpo,lfh}. Task-matched demonstrations raise the initial success rate but shift the burden back to collecting expert data~\cite{crsfd}. Learned visual or vision–language evaluators~\cite{rlvlmf,vlm_success} provide denser feedback, but require additional pipelines for data preparation, training, calibration, and deployment~\cite{rapl}. We therefore ask whether a pretrained policy can obtain dense, task-relevant feedback from its predictive models, without training an additional task-specific reward model or requiring online robot interaction.

\begin{figure}[t]
\vspace{6pt}
    \centering
    \includegraphics[width=0.8\columnwidth]{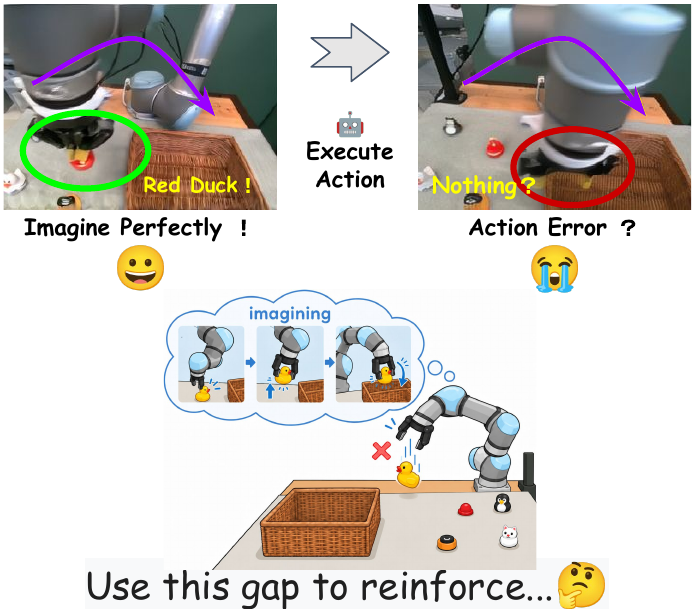}
    \caption{\textbf{Imagination–consequence gap}. From the same instruction and initial state, the world–action model imagines a trajectory that completes the task, but its generated actions fail to execute it. The model already knows what success looks like before it can act on it — we turn this gap into a training signal.}
    \label{fig:intro-gap}
    \vspace{-15pt}
\end{figure}

A WAM provides a source of task-level visual prediction, since its generated video can depict possible task-completion outcomes. During training, however, the two output streams can diverge: the generated video may show task completion, while the action chunk generated alongside it may not lead to the corresponding predicted consequence (Fig.~\ref{fig:intro-gap}). The model can develop stronger task-completion visual prediction before generating action sequences that reliably lead to such outcomes, suggesting an opportunity to improve action generation. However, the world model's visual predictions (videos) do not embed physics information or provide feasibility validation, indicating that a visual plan is not necessarily a valid motor plan. Under these constraints, we treat the WAM-generated visual prediction as a goal-conditioned visual proposal and use a separately pretrained, frozen action-conditioned world model to estimate the consequences of generated actions. The feedback signal measures trajectory-level agreement between the two predictions and alignment of the final predicted frame with the task goal. Because the video head and consequence model are frozen during optimization, the policy cannot directly modify the predictive models that generate the feedback signal. This provides dense rollout-level feedback without online robot interaction, reward labels, or a separately trained success detector, which can then be used for policy optimization.

\begin{figure*}[t]
    \centering
    \vspace{6pt}
    \includegraphics[width=\textwidth]{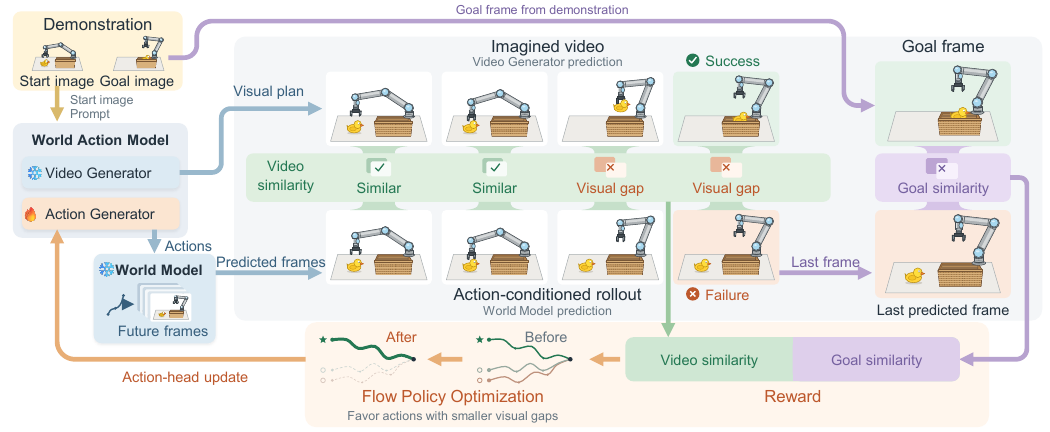}
    \vspace{-20pt}
   \caption{\textbf{Overview of imagination-action alignment for robot manipulation}. From the start image and instruction, WAM rolls out future video and actions. A frozen action-conditioned world model predicts what those actions would actually produce and supplies the next observation. We evaluate action consequences using trajectory similarity between the two videos and goal similarity against a demonstration goal frame. Because both video predictors stay frozen (snowflake), the policy can only raise these rewards by changing its actions. These rewards benefit FPO, which updates the action head (flame) using a clipped surrogate objective.}
\vspace{-15pt}
    \label{fig:framework}
    
\end{figure*}

That step is not straightforward for a flow-based WAM. Existing RL methods for flow policies often rely on explicit sampling formulations or stochastic reformulations to obtain tractable optimization objectives~\cite{pi_rl,flow_grpo}. Cosmos3 instead samples video and actions using UniPC, a high-order multi-step solver for which such per-step likelihood formulations are not directly available. We therefore adapt Flow Policy Optimization~\cite{fpo}, which builds an advantage-weighted surrogate from conditional flow-matching losses while treating the sampler as a black box. This allows us to optimize a flow-based robot policy under a pretrained high-order multi-step sampler, enabling optimization through Cosmos3’s native UniPC sampling procedure while updating only the action head.

The main contributions of this work are:
\begin{itemize}
  \item We introduce a world-model-guided RL framework that exploits the discrepancy between WAM-generated visual predictions and action-conditioned consequences predicted by a frozen world model to refine robot policies without online robot interaction.
  \item We identify an imagination–consequence gap in world-action models and leverage it as dense rollout-level feedback. A frozen pretrained action-conditioned world model provides feedback while avoiding direct updates to the feedback-generating models.
  \item We adapt Flow Policy Optimization for flow-based world-action policies with pretrained high-order multi-step samplers, enabling action-head-only optimization under the native UniPC sampling procedure.
\end{itemize}

\section{Related Work}
\label{sec:related}
\noindent\textbf{Video and world action models for control:}
Video and world models support robot control by predicting future observations, states, or task-relevant representations for planning and manipulation. One line generates future videos as goal-conditioned visual predictions and maps them into actions with a separate inverse dynamics model~\cite{unipi,vera,robodreamer}, decoupling high-level prediction from low-level control but introducing the challenge of maintaining consistency between predicted visual trajectories and generated actions. A second line predicts observations and actions jointly in a unified model, sharing representations or parameters across both modalities~\cite{gr1,cosmos_policy,dreamzero,fast_wam}. A third line predicts compact latent world states~\cite{vjepa2,lawam,dino_wm,vla_jepa}, improving efficiency while providing less direct access to pixel-level observations for visual consistency evaluation. We follow the joint world-action modeling paradigm and build on Cosmos3~\cite{cosmos3}. However, existing joint models optimize visual and action generation objectives separately, while explicit alignment between predicted visual outcomes and action-conditioned consequences remains insufficiently studied.

\noindent\textbf{World models for action outcome prediction:}
Action-conditioned world models predict future observations conditioned on states and actions, estimating possible action consequences. Existing approaches either use explicit action conditioning with control semantics~\cite{ctrl_world,ivideogpt,unisim,irasim} or learn latent action representations from video~\cite{genie,adaworld,playable_environments}. While latent-action models enable action-free learning and cross-embodiment transfer, their latent interfaces are not directly aligned with explicit robot commands. Ctrl-World follows the explicit-action paradigm and supports long-horizon prediction through multi-view prediction, action conditioning, and pose-based memory retrieval~\cite{ctrl_world}. We use a frozen pretrained action-conditioned world model as an external consequence predictor, whose rollouts provide alignment feedback between action consequences and WAM visual predictions.

\noindent\textbf{RL for generative robot policies:}
Reinforcement learning for generative robot policies optimizes pretrained diffusion or flow-based policies using task-level feedback while preserving their sampling procedures. DDPO optimizes diffusion policies with policy gradients~\cite{ddpo}, $\pi$RL extends flow-based RL to VLA models~\cite{pi_rl}, and Flow-GRPO introduces SDE-based formulations for flow optimization~\cite{flow_grpo}. However, these methods typically rely on explicit sampling formulations that are not directly available for the high-order solvers used by recent world-action models. FPO addresses this limitation by deriving an advantage-weighted surrogate from conditional flow-matching losses while treating the sampler as a black box~\cite{fpo}. Unlike WAM-RL, which relies on environment interaction and jointly updates the actor and world model~\cite{wam_rl}, we perform offline post-training with frozen predictive models and optimize only the action head using FPO under Cosmos3’s native UniPC sampler.

\section{Preliminaries}
\label{sec:preliminaries}
\subsection{Flow-matching generative policies}
Flow matching trains continuous normalizing flows by regressing a
time-dependent vector field along prescribed conditional probability paths
between noise and data. Given a condition $c$, the learned
field $v_\theta$ defines the ordinary differential equation
\begin{equation}
    z_{0}\sim p_{0},
    \qquad
    \frac{\mathrm{d}z_{\tau}}{\mathrm{d}\tau}
    =
    v_{\theta}\!\left(z_{\tau},\tau\mid c\right),
    \qquad
    \tau\in[0,1],
    \label{eq:flow-ode}
\end{equation}
where $p_0$ is the base distribution and $z_\tau$ the state at time $\tau$
along the trajectory integrated from $z_0$, with $\tau=0$ and $1$ corresponding
to noise and the generated endpoint.

\subsection{Flow Policy Optimization}
Flow Policy Optimization (FPO) constructs a surrogate policy ratio from conditional flow-matching (CFM) losses, avoiding the
need to evaluate the exact likelihood of a flow-based policy. For a generated sample $z$ under condition $c$, let $\widehat L_{\mathrm{CFM},\theta}(z\mid c)$ denote an empirical estimate of its CFM loss under parameters $\theta$. FPO defines the surrogate ratio as
\begin{equation}
    \widehat{\rho}^{\mathrm{FPO}}_{\theta}(z\mid c)
    =
    \exp\!\left[
        \widehat L_{\mathrm{CFM},\theta_{\mathrm{old}}}(z\mid c)
        -
        \widehat L_{\mathrm{CFM},\theta}(z\mid c)
    \right],
    \label{eq:fpo-ratio}
\end{equation}
where $\theta_{\mathrm{old}}$ generated $z$ and $\theta$ is the current policy
being optimized. Both losses use the same sample, condition, and time--noise
pairs. This ratio is a surrogate for policy change, not the exact likelihood
ratio $p_\theta(z\mid c)/p_{\theta_{\mathrm{old}}}(z\mid c)$.

\section{Method}
\label{sec:method}

Fig.~\ref{fig:framework} overviews our world-model-guided reinforcement-learning framework. The world-action model generates a goal-conditioned visual prediction and an action chunk; a frozen action-conditioned world model predicts the visual consequences of that action chunk and supplies the next observation. The discrepancy between the two predictions, together with alignment to a terminal visual goal, forms the feedback signal used to optimize the action head, avoiding additional real-robot interaction or training of task-specific reward models.

\subsection{Problem Formulation}
\label{sec:problem-formulation}
At rollout step $k$, the world-action model samples a latent endpoint $z_k$ conditioned on the current observation $o_k$ and instruction $g$. Denote its distribution as $\Pi_\theta^z$ and the decoding mapping as $D$, which outputs the generated visual prediction $V_k^C$ and action chunk $a_k\in\mathbb{R}^{H_a\times d_a}$:
\begin{equation}
\begin{aligned}
    z_k
    &\sim
    \Pi_{\theta}^{z}
    \!\left(\cdot\mid o_k,g\right),
    \left(V_k^{C},a_k\right)
    &=
    D\!\left(z_k\right),
\end{aligned}
\label{eq:wam-output}
\end{equation}
The action chunk is converted into the control representation required by the pretrained frozen action-conditioned world model $W_{\phi^0}$, which predicts the action-conditioned visual consequence.
\begin{equation}
    V_k^{W}
    =
    W_{\phi^0}
    \!\left(
        o_k,T_a(a_k)
    \right),
    \label{eq:world-consequence}
\end{equation}
where $T_a$ denotes the action conversion and temporal resampling interface. The two visual predictions are complementary: $V_k^{C}$ is conditioned on the task goal, $V_k^{W}$ on the proposed actions.

We partition the world-action model parameters into the frozen video-generation parameters $\theta_v^0$ and the trainable action-related parameters $\theta_a$, with $\theta=(\theta_v^0,\theta_a)$. Policy improvement does not require real-robot transitions or task-success labels; the framework instead constructs a model-mediated trajectory
\[
X=
\{(o_k,a_k,V_k^C,V_k^W)\}_{k=0}^{N-1}
\] and optimizes
\begin{equation}
    \theta_a^{\star}
    =
    \arg\max_{\theta_a}
    \mathbb{E}_{X
    \sim
    p_{\theta_a}
    (X\mid o_0,g;\theta_v^0,\phi^0)}
    \left[
        \sum_{k=0}^{N-1}
        \gamma^k r_k
    \right],
    \label{eq:framework-objective}
\end{equation}
with $\theta_v^{0}$ and $\phi^{0}$ fixed throughout training. The reward $r_k$ and transition to $o_{k+1}$ are defined in \eqref{eq:combined_reward} and \eqref{eq:model-transition}.

\subsection{Method Overview}
The core idea is to use predicted action consequences as feedback, encouraging actions whose predicted outcomes align with the model’s goal-conditioned visual predictions. At each chunk, the WAM generates a goal-conditioned visual prediction and action sequence, while a frozen action-conditioned world model predicts their consequences and provides the next observation. The feedback combines trajectory-level consistency with terminal visual-goal alignment. We optimize only the action head using an FPO surrogate with group-relative advantages, while keeping the visual path and world model frozen. Algorithm~\ref{alg:loop} summarizes the training procedure.

\subsection{Closed-Loop Cross-Model Rollout}
\label{sec:closed-loop-rollout}

We instantiate the world action model $\Pi_\theta$ with a native Cosmos3 policy~\cite{cosmos3} and the action-conditioned world model $W_{\phi^0}$ with a frozen Ctrl-World model~\cite{ctrl_world}.

For each policy step, Cosmos3 jointly generates the goal-conditioned visual prediction $V_k^{C}$ and action chunk $a_k$ in \eqref{eq:wam-output}. Before passing to Ctrl-World, $T_a$ restores the physical scale of the actions, converts the policy representation into the world-model control interface, and resamples the commands to the world-model rate. Ctrl-World then generates the consequence video $V_k^{W}$ from the current observation and the converted actions.

The rollout is closed by using the predicted consequence to construct the observation for the next policy step:
\begin{equation}
    o_{k+1}
    =
    \operatorname{Ext}_{\mathrm{obs}}
    \!\left(
        V_k^{W}
    \right),
    \label{eq:model-transition}
\end{equation}
where $\operatorname{Ext}_{\mathrm{obs}}$ extracts the latest multi-view observation after the action chunk. The world model retains image and action histories within each rollout and updates them after each chunk; we omit these history inputs for brevity. Importantly, $V_k^{C}$ serves only as the target for cross-model feedback and is never fed back as the next observation. Repeating \eqref{eq:wam-output}--\eqref{eq:model-transition} for $N$ chunks yields a closed-loop trajectory in which subsequent actions are conditioned on the predicted consequences of earlier actions.

For group-relative optimization, we sample $B$ groups, each containing $G$ trajectories that share the same reset observation and instruction:
\begin{equation}
\begin{aligned}
o_{b,i,0}&=o_{b,0},\qquad g_{b,i}=g_b,\\
b&=1,\ldots,B,\qquad i=1,\ldots,G,
\end{aligned}
\label{eq:shared-reset}
\end{equation}
Within each group, independent draws from $p_0$ yield different latent endpoints and hence different imagined futures and action proposals under the same initial task condition.

\begin{algorithm}[h]
\caption{Closed-loop post-training}
\label{alg:loop}
\begin{algorithmic}[1]
\Require policy $\Pi_{\theta_a}$, frozen world model $W$,
reset set $D_{\mathrm{reset}}$, goal image $I_g^\star$,
group size $G$, horizon $N$, discount $\gamma$

\State $\theta_{a,\mathrm{old}}\gets\theta_a$;
sample $(o_0,g)\sim D_{\mathrm{reset}}$
\For{$i=1,\ldots,G$}
    \State $o_{i,0}\gets o_0$
    \For{$k=0,\ldots,N-1$}
        \State $(V^C_{i,k},a_{i,k},z_{i,k})
        \gets\operatorname{NativeUniPC}
        (\Pi_{\theta_{a,\mathrm{old}}},o_{i,k},g)$
        \State $V^W_{i,k}\gets W(o_{i,k},T_a(a_{i,k}))$
        \State Compute $r_{i,k}$
        \State Store $z_{i,k}$, $c_{i,k}=(o_{i,k},g)$,
        and fixed probes $\Omega_{i,k}$
        \State $o_{i,k+1}\gets
        \operatorname{Ext}_{\mathrm{obs}}(V^W_{i,k})$
    \EndFor
\EndFor

\State Compute $R_{i,k}$ and $\widehat A_{i,k}$
\For{each minibatch of stored chunks}
    \For{each chunk $(i,k)$ in the minibatch}
        \State Compute $\widehat L^a_{\mathrm{old}}$ and
        $\widehat L^a$
        \State $\widehat\rho_{i,k}\gets
        \exp(\widehat L^a_{\mathrm{old}}-\widehat L^a)$
    \EndFor
    \State Update $\theta_a$ once using the minibatch average
\EndFor
\end{algorithmic}
\end{algorithm}
 
\subsection{Cross-Model Feedback and Temporal Credit}
\label{sec:cross-model-feedback}

We construct feedback from trajectory consistency and terminal-goal alignment. For trajectory $i$ in group $b$ at chunk $k$, the trajectory feedback measures the discrepancy between the WAM-generated visual prediction and the action-conditioned prediction. We select frames at $H_v$ common timestamps from both videos and compare corresponding views at matching resolution. For vectorized frames $F,F'$, we use pixel MSE, $D_{\mathrm{traj}}(F,F')=D_{\mathrm{goal}}(F,F')=\|F-F'\|_2^2/P$, where $P$ counts all compared pixel values across channels and views, and pixel values are on the $[0,255]$ scale. The trajectory reward is

\begin{equation}
    r^{\mathrm{traj}}_{b,i,k}
    =
    -\frac{1}{H_v}
    \sum_{t=1}^{H_v}
    D_{\mathrm{traj}}\left(
        V^C_{b,i,k,t},
        V^W_{b,i,k,t}
    \right).
    \label{eq:trajectory_reward}
\end{equation}

To encourage task completion, we additionally compare $V_{\mathrm{final}}^W$, the final frame predicted by the world model, with a predefined visual goal reference $I_g^\star$ for each task $g$, preprocessed identically to the predicted frames. Omitting group and trajectory indices in
\eqref{eq:goal_reward}--\eqref{eq:combined_reward}, the goal reward is
\begin{equation}
r^{\mathrm{goal}}
=-D_{\mathrm{goal}}\!\left(V_{\mathrm{final}}^W,I_g^\star\right).
\label{eq:goal_reward}
\end{equation}
The reward used for policy optimization is
\begin{equation}
r_k=\lambda_{\mathrm{traj}}r_k^{\mathrm{traj}}
+\lambda_{\mathrm{goal}}r_k^{\mathrm{goal}}.
\label{eq:combined_reward}
\end{equation}
where the nonnegative weights $\lambda_{\mathrm{traj}}$ and $\lambda_{\mathrm{goal}}$ balance the two terms, and the goal term equals \eqref{eq:goal_reward} at the final chunk and is zero otherwise. Higher feedback scores indicate smaller pixel discrepancies: the trajectory term encourages agreement with the WAM's imagined motion, the goal term encourages progress toward the desired outcome.

A trajectory contains $N$ consecutive action chunks. To assign an earlier chunk credit for the consequences predicted at later rollout steps, we compute the discounted suffix return
\begin{equation}
    R_{b,i,k}
    =
    \sum_{j=k}^{N-1}
    \gamma^{j-k} r_{b,i,j}.
    \label{eq:suffix-return}
\end{equation}
Returns are normalized separately at each chunk position across the $G$
trajectories sharing the same reset state:
\begin{equation}
\widehat A_{b,i,k}
=\frac{R_{b,i,k}-\mu_{b,k}}{s_{b,k}+\delta},
\label{eq:group-advantage}
\end{equation}
where $\mu_{b,k}$ and $s_{b,k}$ are the mean and population standard deviation of the $G$ returns within group $b$ at chunk $k$, and $\delta>0$ ensures numerical stability.

\begin{figure*}[t]
    \centering
    \includegraphics[width=\textwidth]
    {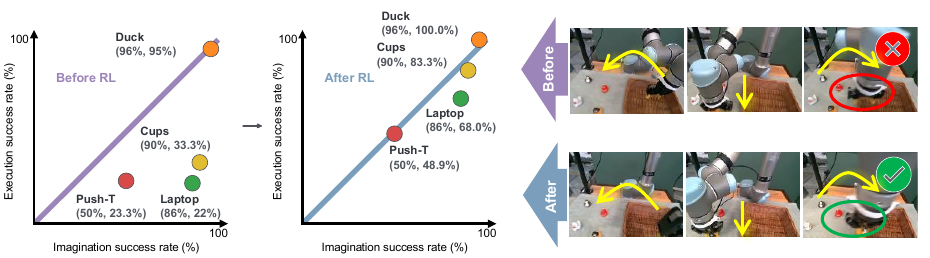}
    \vspace{-15pt}
    \caption{\textbf{Imagination and execution success rates before (left) and after (right) RL}. Each point is one task; the diagonal marks perfect agreement, so distance below it is the imagination–consequence gap. The gap between imagination and execution success rates narrows after post-training. Far right: a duck grasp that fails before RL and succeeds after.}
    \label{fig:exp}
    \vspace{-15pt}
\end{figure*}

\subsection{Action-Head Policy Optimization}
\label{sec:action_fpo}
The group-relative advantages defined above provide the feedback for updating the WAM's action head. We use them in a clipped FPO objective whose surrogate ratio is computed from action-head CFM losses, connecting the cross-model feedback to policy optimization without requiring gradients through the reward computation or the action-conditioned world model.

For member $i$ of group $b$ at chunk $k$, the WAM generates a joint latent endpoint $z_{b,i,k}=[z^v_{b,i,k};z^a_{b,i,k}]$ using its native UniPC sampler, which we evaluate with the FPO surrogate introduced in the preliminaries~\cite{fpo}. This retains the native rollout procedure without requiring a likelihood over the numerical solver trajectory.

For each stored endpoint, we sample $Q$ time--noise pairs $\Omega_{b,i,k}=\{(\tau_q,\boldsymbol{\epsilon}_q)\}_{q=1}^{Q}$, where $\tau_q\sim\mathrm{U}(0,1)$ and $\boldsymbol\epsilon_q\sim\mathrm{N}(0,I)$ are independent and shared between the behavior and current policies. Following the noise-to-data convention, the perturbed input is
\begin{equation}
z^{(q)}_{b,i,k}
=z_{b,i,k}+(1-\tau_q)M_{b,i,k}\odot
\left(\boldsymbol\epsilon_q-z_{b,i,k}\right),
\label{eq:fpo_joint_renoise}
\end{equation}
The mask $M_{b,i,k}$ selects future visual latents and valid action coordinates, leaving conditioning and padding unchanged.

We compute the CFM loss only on valid action coordinates:
\begin{equation}
\widehat L^a_{\theta_a,b,i,k}
=\frac1Q\sum_{q=1}^{Q}
\left\|P_a\!\left[
v_{\theta_v^0,\theta_a}
\left(z^{(q)}_{b,i,k},\tau_q\mid c_{b,i,k}\right)
-u^{(q)}_{b,i,k}\right]\right\|_2^2,
\label{eq:action_cfm_loss}
\end{equation}
where $u^{(q)}_{b,i,k}=M_{b,i,k}\odot(z_{b,i,k}-\boldsymbol\epsilon_q)$ is the target velocity of \eqref{eq:fpo_joint_renoise}, $P_a$ selects valid action coordinates, and $c_{b,i,k}=(o_{b,i,k},g_b)$. This loss defines the surrogate ratio for updating the action head:
\begin{equation}
\widehat{\rho}^{\mathrm{FPO}}_{b,i,k}(\theta_a)
=
\exp\!\left(
\widehat{L}^{a}_{\theta_{a,\mathrm{old}},b,i,k}
-
\widehat{L}^{a}_{\theta_a,b,i,k}
\right),
\label{eq:action_fpo_ratio}
\end{equation}
where both losses use the same endpoint, condition, and time--noise pairs.

We combine this ratio with the group-relative advantage $\widehat{A}_{b,i,k}$ and maximize the critic-free clipped objective:
\begin{equation}
J_{\mathrm{FPO}}(\theta_a)
=\mathbb E\!\left[
\min\!\left(
\rho\widehat A,\,
\operatorname{clip}(\rho,1-\epsilon_{\mathrm{clip}},1+\epsilon_{\mathrm{clip}})
\widehat A\right)\right],
\label{eq:clipped_fpo}
\end{equation}
where $\rho$ and $\widehat A$ denote the ratio in \eqref{eq:action_fpo_ratio} and the advantage in \eqref{eq:group-advantage}, the expectation averages uniformly over the $BGN$ collected chunks, and $\epsilon_{\mathrm{clip}}\in(0,1)$ is the clipping threshold. The sampled trajectories, probe pairs, advantages, and behavior-policy losses remain fixed during optimization. Only the action-head parameters $\theta_a$ are updated; the WAM’s visual-generation parameters and the action-conditioned world model stay frozen, preventing direct updates to the predictive models during optimization.

\section{Experiments}

We evaluate the proposed framework through four questions.
\begin{itemize}[
  \setlength{\topsep}{0pt}%
  \setlength{\itemsep}{-2pt}%
  \nocalcleftmargintrue
  \setlength{\leftmargin}{1.2em}%
]
    \item \textbf{Q1:} How consistent are the visual predictions of a world–action model with the action-conditioned consequences predicted from its generated actions?
\item \textbf{Q2:} Can policy optimization guided by this model-derived feedback improve real-world robot performance?
\item \textbf{Q3:} How do trajectory-level and terminal-goal feedback components affect policy optimization?
\item \textbf{Q4:} How do model-derived feedback signals relate to predicted task success?
\end{itemize}
To address these questions, we focus on four aspects: model consistency, real-robot performance, feedback mechanisms, and reward effectiveness.

\begin{figure}[t]
    \centering
    \includegraphics[width=\linewidth]{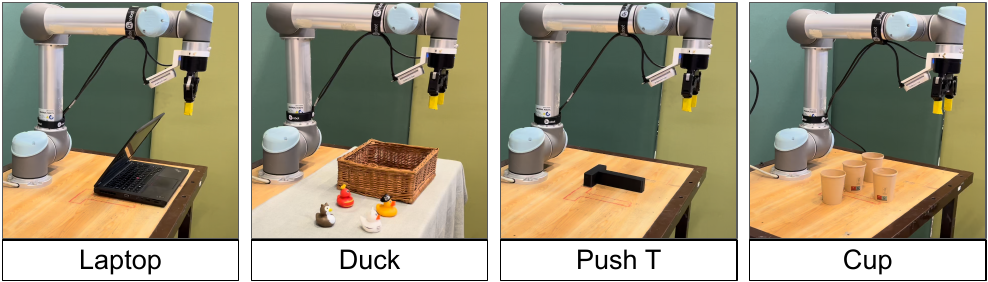}
    \caption{\textbf{Experimental tasks}: closing a laptop, placing a specified colored duck into a basket (four colors, selected from the instruction), pushing a T-shaped block to a target pose, and nesting four cups into one stack. We collect 40, 200, 150, and 100 demonstrations for the respective tasks; Cup has the longest horizon.}
    \label{fig:task_demonstration}
    \vspace{-10pt}
\end{figure}

\subsection{Experimental Setup}

\subsubsection{Tasks}
As shown in Fig.~\ref{fig:task_demonstration}, we evaluate four real-world
manipulation tasks on a UR5 robot:
\begin{itemize}[
  \setlength{\topsep}{0pt}%
  \setlength{\itemsep}{-2pt}%
  \nocalcleftmargintrue
  \setlength{\leftmargin}{1.2em}%
]
  \item \emph{Laptop closing}: The robot rotates the open lid until the laptop is fully shut without displacing the base (40 demonstrations).
  \item \emph{Duck placing}: The robot grasps the colored duck specified by the instruction from four candidates on the table and places it into a basket. All four colors appear during training (200 demonstrations, 50 per color).
  \item \emph{Push-T}: The robot translates and reorients a T-shaped block to match a target pose (150 demonstrations).
  \item \emph{Cup nesting}: The robot stacks four cups into a single column, requiring four successive grasp-and-place motions without disturbing the partial stack (100 demonstrations).
\end{itemize}
Laptop closing and cup nesting require accurate contact establishment and maintenance during execution, while Push-T and duck placing require visual inference of object pose and identity.

\subsubsection{Compared methods}
We compare the full framework against its SFT initialization and three external systems: Fast-WAM~\cite{fast_wam}, VLA-JEPA~\cite{vla_jepa} and $\pi_{0.5}$~\cite{pi05}. Each retains its official model, training procedure, observation and action interface, and default execution horizon, while being adapted and evaluated using the same demonstration data and task protocol.

\subsubsection{Training and checkpoint selection}
All systems are adapted from the same task-specific UR5 demonstration split, divided into training and testing sets at a 4:1 ratio. For each method, we select the checkpoint with the lowest validation loss during adaptation.

\subsubsection{Implementation details}
We combine FPO with group-relative advantages. Each rollout batch contains $B=8$ groups of $G=16$ trajectories that share the same initial observation and instruction and differ only through independent policy sampling noise. Each trajectory contains $N=5$ action chunks, and discounted suffix returns are normalized within each group at each chunk position with $\gamma=0.95$.

\begin{figure}
    \centering
    \vspace{6pt}    \includegraphics[width=0.75\columnwidth]
    {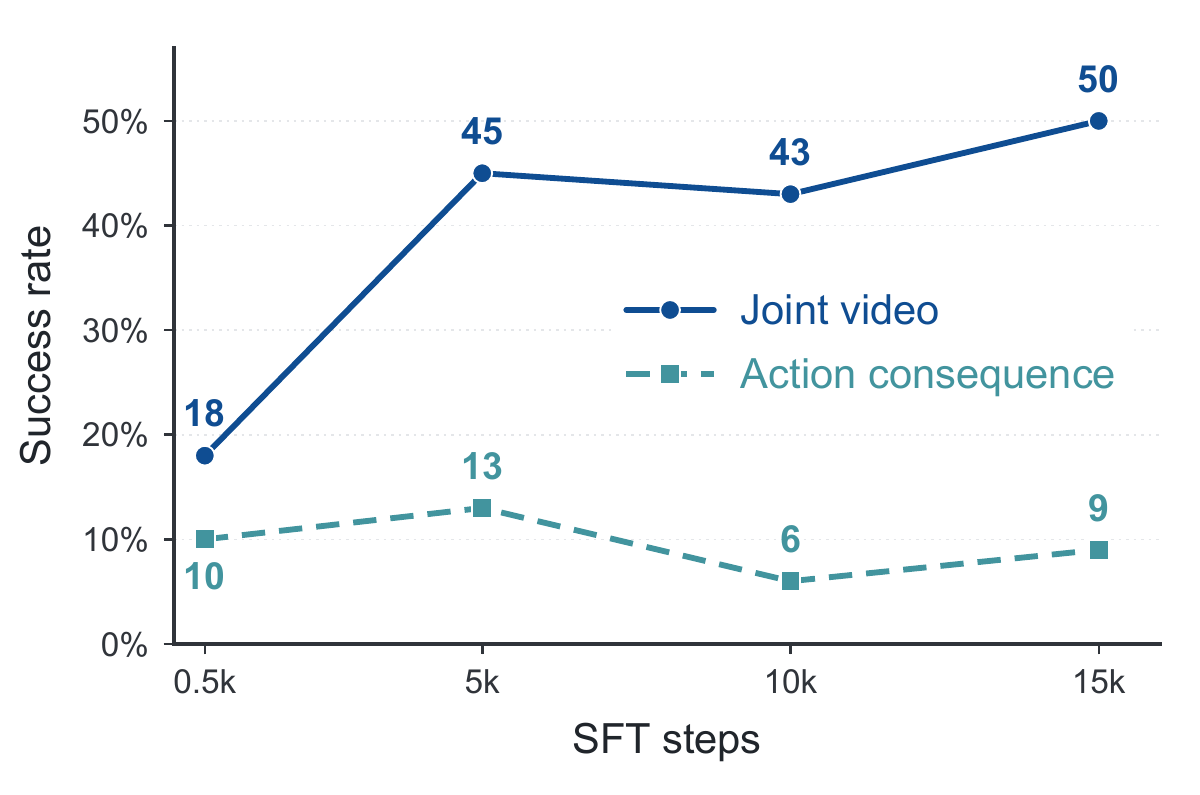}
     \vspace{-5pt}
    \caption{\textbf{Model-space success during SFT.} Both streams are scored by
    a frozen classifier on the terminal frame. The imagined video increasingly
    shows task completion ($18\to50\%$) while the world model's rollout of the
    same actions does not (6--13\%), leaving a 41-point gap at 15k steps.}
    \label{fig:sft_gap}
    \vspace{-15pt}
\end{figure}

\subsection{Model-Space Video--Consequence Gap During SFT}
\label{sec:gap}

We first test whether task-completion predictions from the WAM visual stream match consequences predicted from generated actions. At four SFT checkpoints (500, 5k, 10k, and 15k steps), we sample 100 paired rollouts from 20 initializations. Within each pair, the video stream recursively conditions on its predictions while the frozen world model predicts the consequences of the same action chunks from the same initial observation, and a frozen instruction-conditioned classifier scores the terminal predictions.

Fig.~\ref{fig:sft_gap} shows the resulting model-space success scores. As SFT proceeds, the WAM visual predictions increasingly depict task completion, rising from 18\% to 50\%, while the corresponding action-conditioned predictions remain at 6–13\% throughout. The paired gap widens to 41 percentage points at 15k steps. These results indicate an asymmetry between visual prediction and action generation: the model can produce task-completion visual predictions before generating action sequences whose predicted consequences achieve similar outcomes. Because both scores use the same classifier and initial states, this gap reflects inconsistency between two model-generated predictions rather than differences in the evaluation protocol, and it motivates the feedback defined in Eq.~\eqref{eq:combined_reward}.

\begin{figure*}[t]
    \centering
    \includegraphics[width=\textwidth]
    {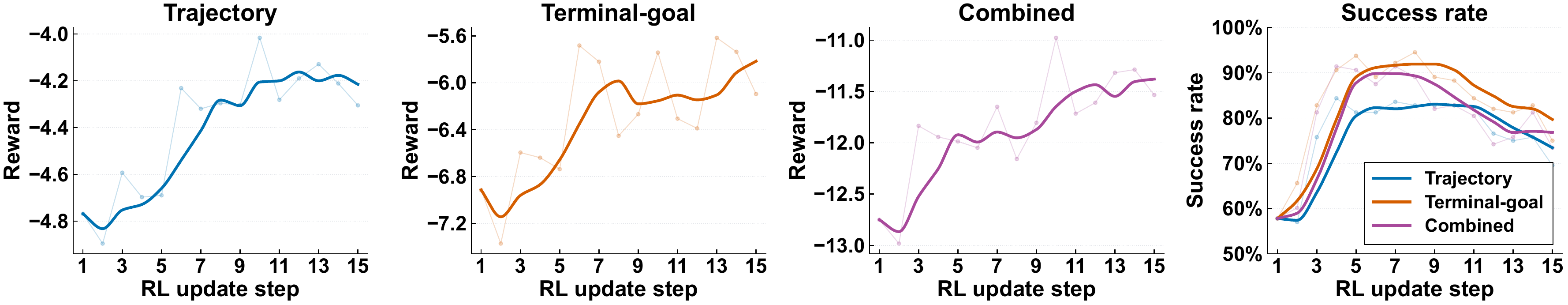}
 
    \caption{\textbf{Reward ablation on the close-laptop task, varying only the reward composition}. Left three: Trajectory, Terminal-goal, and Combined rewards over FPO updates, each rising under its own objective. Right: classifier-estimated model-space success, which improves early then diverges — model-space success does not directly track real-robot performance.}
    \label{fig:reward_ablation}
    \vspace{-15pt}
\end{figure*}

\subsection{Overall Real-Robot Performance}

Table~\ref{tab:main_results} reports real-robot success rates, averaged over four tasks with equal weight. After RL, our model achieves the highest average success rate among evaluated methods at 75.1\%, improving over its SFT initialization (43.4\%) by 31.7 percentage points and over the strongest baseline, $\pi_{0.5}$, by 13.7 percentage points. The results show that policy optimization with world-model feedback provides gains beyond supervised imitation alone.

\begin{table}[h]
\centering
\caption{Real-robot task success rates (\%).}
\vspace{-5pt}
\label{tab:main_results}
\renewcommand{\arraystretch}{0.85}
\setlength{\tabcolsep}{3pt}
\small

\begin{tabular*}{\columnwidth}{@{\extracolsep{\fill}}lccccc@{}}
\toprule
Method & Laptop & Duck & Push-T & Cup & Avg. \\
\midrule
Ours (SFT)  & 22.0\%   & 95.0\%  & 23.3\% & 33.3\% & 43.4\% \\
Fast-WAM    & 16.0\%   & 0.0\%     & 0.0\%    & 0.0\%    & 4.0\% \\
VLA-JEPA    & 15.0\%   & 20.0\%     & 0.0\%    & 0.0\%    & 8.8\% \\
$\pi_{0.5}$ & 56.7\% & 85.0\%  & 33.3\% & 70.6\% & 61.4\% \\
\midrule
\textbf{Ours (RL)} & \textbf{68.0\%}
            & \textbf{100.0\%}
            & \textbf{48.9\%}
            & \textbf{83.3\%}
            & \textbf{75.1\%} \\
\bottomrule
\end{tabular*}
\vspace{-10pt}
\end{table}

The improvement is consistent across all four tasks: 22.0\% to 68.0\% on laptop, 95.0\% to 100.0\% on duck, 23.3\% to 48.9\% on Push-T, and 33.3\% to 83.3\% on cup. Compared with $\pi_{0.5}$, our method achieves higher success rates on all evaluated tasks, with gains of 11.3, 15.0, 15.6, and 12.7 percentage points, respectively. Under the same adaptation protocol, VLA-JEPA and Fast-WAM achieve lower success rates than our SFT initialization. Both were developed with substantially larger adaptation budgets than the 40--200 demonstrations per task available here, so we read these numbers as evidence that the low-data regime is difficult rather than as an inherent limitation of either method.

Table~\ref{tab:execution_efficiency} shows that execution efficiency also improves after RL. Mean completion time and mean control step count decrease on all four tasks: by 10.2\% and 9.9\% on laptop, 10.7\% and 14.7\% on duck, 6.0\% and 3.8\% on Push-T, and 16.8\% and 25.1\% on the long-horizon cup task. The medians follow the same trend except on duck, where they are essentially unchanged, indicating that the mean reduction there comes from fewer long, hesitant trials rather than a uniformly faster policy. These statistics are computed over each policy's successful trials, so the two rows cover different trial sets; the consistent direction across tasks nonetheless suggests that improving consistency between generated actions and predicted visual outcomes reduces corrective motion rather than trading accuracy for speed.

\begin{table}[h]
\centering
\caption{Real-robot task execution efficiency.}
\vspace{-5pt}
\label{tab:execution_efficiency}
\renewcommand{\arraystretch}{0.85}
\setlength{\tabcolsep}{3pt}
\small
\begin{tabular*}{\columnwidth}{@{\extracolsep{\fill}}lcccccc@{}}
\toprule
Task & Method & $n$
& \multicolumn{2}{c}{Time (s) $\downarrow$}
& \multicolumn{2}{c}{Control steps $\downarrow$} \\
\cmidrule(lr){4-5}\cmidrule(lr){6-7}
& & & Mean & Median & Mean & Median \\
\midrule
Laptop   & SFT & 11 & 46.1 & 40.8 & 123.6 & 113 \\
       & RL & 34 & \textbf{41.4} & \textbf{39.7} & \textbf{111.4} & \textbf{107} \\
\midrule
Duck   & SFT & 57 & 63.5 & \textbf{54.1} & 200.0 & \textbf{160} \\
       & RL & 60 & \textbf{56.7} & 54.2 & \textbf{170.7} & \textbf{160} \\
\midrule
Push-T & SFT & 21 & 131.6 & 124.4 & 289.3 & 279 \\
       & RL & 44 & \textbf{123.7} & \textbf{121.9} & \textbf{278.3} & \textbf{272} \\
\midrule
Cup   & SFT & 10 & 206.0 & 196.7 & 667.6 & 632 \\
       & RL & 25 & \textbf{171.3} & \textbf{173.4} & \textbf{500.0} & \textbf{496} \\

\bottomrule
\end{tabular*}
\par\smallskip
\begin{minipage}{\columnwidth}
\footnotesize
Statistics use each policy's successful trials; $n$ is the number of
successes. Time includes inference
waiting. Control steps count executed actions.
\end{minipage}
\vspace{-15pt}
\end{table}

\subsection{Ablation Study of Reward}

We evaluate the respective contributions of trajectory-level and terminal-goal feedback through a feedback-component ablation on the \textit{close-laptop} task, keeping the SFT initialization, Ctrl-World model, UniPC sampler, FPO objective, and training budget constant and varying only the reward composition. We compare three settings: \textit{Trajectory reward}, which uses only the trajectory-consistency term; \textit{Terminal reward}, which uses only the terminal-goal term; and \textit{Combined reward}, which integrates both.

As shown in the left three panels of Fig.~\ref{fig:reward_ablation}, all three settings improve steadily over FPO updates, each rising under its own objective. Both feedback types therefore provide useful optimization signals for FPO individually. On the right of Fig.~\ref{fig:reward_ablation}, all three settings improve sharply in the early updates, peak around updates 9--11, and then decline. Because this metric comes from a classifier applied to Ctrl-World's predicted terminal states, it measures completion only within the model space, and the late decline may indicate a mismatch between the optimized policy and the frozen world model's prediction distribution rather than genuine degradation. We accordingly use model-space success only to compare optimization signals during training.

These results suggest that the two feedback components play complementary roles: the trajectory feedback constrains consistency during intermediate predicted evolution, whereas the terminal feedback focuses on the final predicted goal state. Combining feedback from both the process and the endpoint yields a
more comprehensive optimization signal.

\begin{figure}[h]
    \centering
\includegraphics[width=\columnwidth]
    {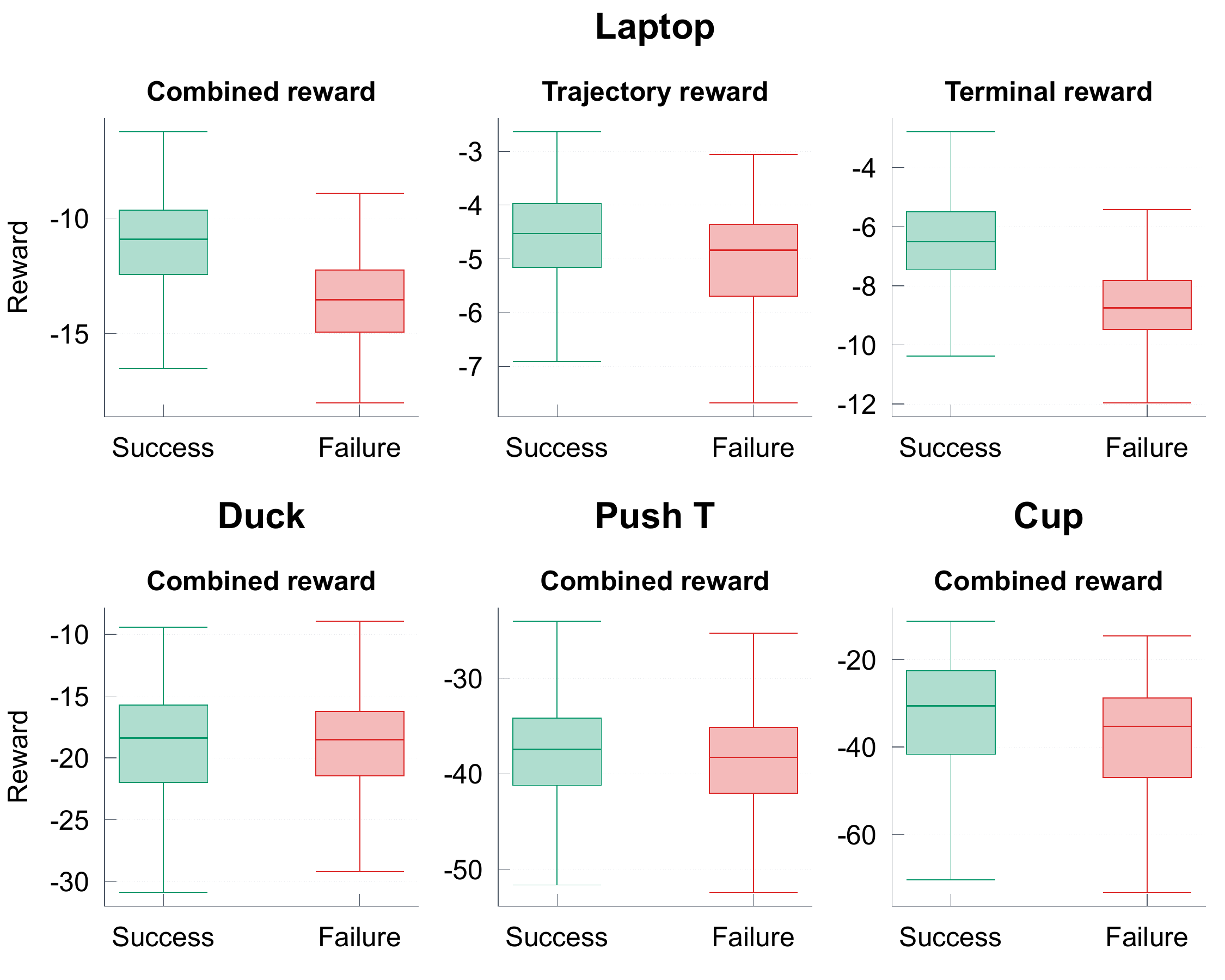}
      \vspace{-15pt}
    \caption{\textbf{Reward distributions for successful (green) and failed (red) rollouts}. Laptop is shown for all three reward settings; other tasks use the combined reward. Successes score higher throughout, but the margin is clear on Laptop and Cup and largely overlapping on Duck and Push-T.}
    \label{fig:reward}
      \vspace{-15pt}
\end{figure}

\subsection{Correlation Between Reward and Predicted Success}

\begin{figure}[t]
    \centering
\includegraphics[width=\columnwidth]
    {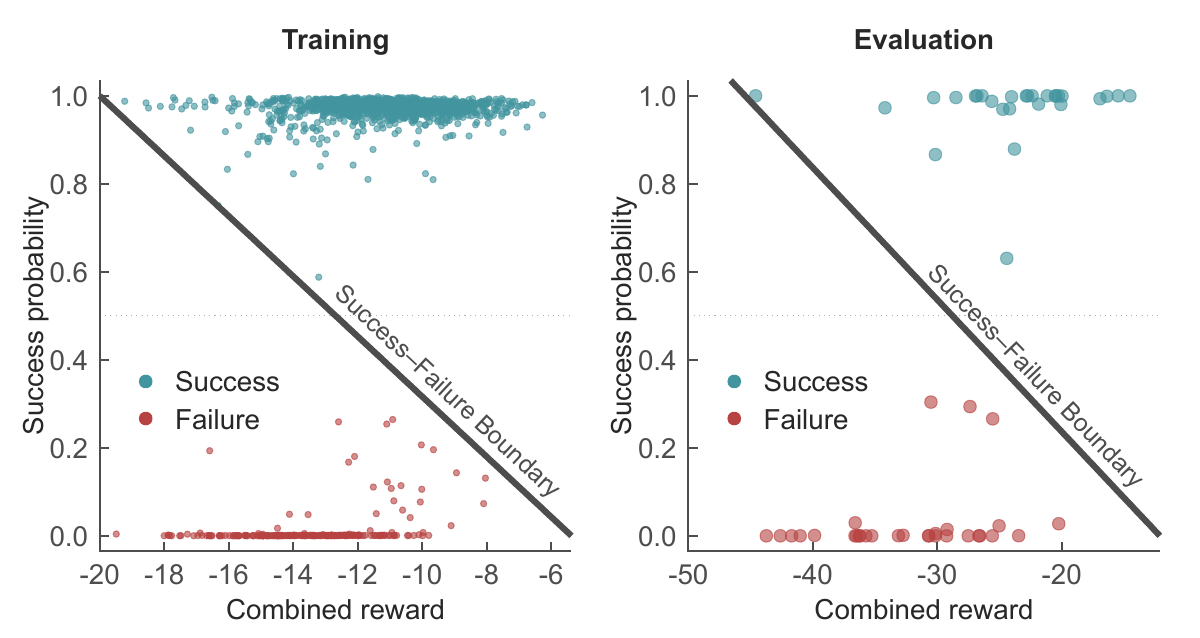}
\vspace{-15pt}
    \caption{\textbf{Combined reward versus predicted success probability on the training (left) and evaluation (right) sets}; each point is one rollout. Successes concentrate at high probability and higher reward, failures at the opposite corner, with similar separation in both splits. Diagonals are visual guides only; both axes come from Ctrl-World predictions, so the correlation is within model space.}
    \label{fig:reward-point}
    \vspace{-15pt}
\end{figure}

We analyze the relationship between the feedback signal and task-prediction success. For each RL rollout, we pair the feedback score with the success classification derived from the final-state image predicted by Ctrl-World, so this analysis reflects the association within the model space. As shown in Fig.~\ref{fig:reward}, this relationship varies across tasks. On laptop and cup, reward distributions for successful rollouts sit clearly above those for failed ones, whereas on duck and Push-T the two overlap substantially. The combined feedback therefore carries directional information consistent with predicted success, though the association strength varies by task. We further examine the relationship at the individual-rollout level. As shown in Fig.~\ref{fig:reward-point}, successful rollouts concentrate at high predicted success probability and at the upper end of the reward range, while failed rollouts fall at the opposite corner, with similar separation in the training and evaluation splits. The combined reward thus orders rollouts consistently with predicted task success, supporting its use as a feedback signal. Because both quantities derive from the same Ctrl-World predictions, this correlation holds within the model space.

Both components show consistent trends: trajectory feedback captures consistency between predicted action evolution and target progression, while terminal feedback emphasizes final-state completion. Combining them supervises both intermediate process and final outcome. Higher feedback scores correspond to more successful model predictions across reward distributions and individual rollouts; although task-dependent, the feedback remains informative for guiding the policy toward improved predicted outcomes.

\section{conclusions}
We propose a policy optimization method that converts the discrepancy between a world-action model’s visual predictions and action-conditioned consequences into an optimization signal for its actions. A frozen pretrained action-conditioned world model predicts the action-conditioned visual consequences of generated actions, and the discrepancy between the WAM visual prediction and the predicted consequence, together with alignment to a terminal visual goal, forms a dense rollout-level feedback signal for policy optimization without directly updating the predictive models. Adapting Flow Policy Optimization to the action head allows this feedback signal to update the policy under the pretrained model’s native UniPC sampler, without online robot interaction or additional task-specific reward-model training. On four real-world UR5 tasks, post-training on a 41-point visual prediction–consequence gap raises the average success rate from 43.4\% to 75.1\%, and ablations confirm the complementary roles of trajectory-level and terminal-goal feedback. Future work will explore world models that explicitly incorporate physical constraints to better assess the feasibility of visual plans and narrow the gap between visual planning and real-world execution, while extending the approach to longer-horizon manipulation.

\bibliography{ref}  

\end{document}